# The European Language Technology Landscape in 2020: Language-Centric and Human-Centric AI for Cross-Cultural Communication in Multilingual Europe

Georg Rehm[1], Katrin Marheinecke[1], Stefanie Hegele[1], Stelios Piperidis[2], Kalina Bontcheva[3], Jan Hajič[4], Khalid Choukri[5], Andrejs Vasiļjevs[6], Gerhard Backfried[7], Christoph Prinz[7], José Manuel Gómez Pérez[8], Luc Meertens[9], Paul Lukowicz[1], Josef van Genabith[1], Andrea Lösch[1], Philipp Slusallek[1], Morten Irgens[10], Patrick Gatellier[11], Joachim Köhler[12], Laure Le Bars[13], Dimitra Anastasiou[14], Albina Auksoriūtė[15], Núria Bel[16], António Branco[17], Gerhard Budin[18], Walter Daelemans[19], Koenraad De Smedt[20], Radovan Garabík[21], Maria Gavriilidou[2], Dagmar Gromann[18], Svetla Koeva[22], Simon Krek[23], Cvetana Krstev[24], Krister Lindén[25], Bernardo Magnini[26], Jan Odijk[27], Maciej Ogrodniczuk[28], Eiríkur Rögnvaldsson[29], Mike Rosner[30], Bolette Sandford Pedersen[31], Inguna Skadiņa[32], Marko Tadić[33], Dan Tufiş[34], Tamás Váradi[35], Kadri Vider[36], Andy Way[37], François Yvon[38]

[1] DFKI GmbH, Germany • [2] ILSP/Athena RC, Greece • [3] University of Sheffield, UK • [4] Charles University, Czech Republic • [5] ELDA, France • [6] Tilde, Latvia • [7] SAIL LABS Technology GmbH, Austria • [8] Expert System Iberia SL, Spain • [9] CrossLang, Belgium • [10] Oslo Metropolitan University, Norway • [11] Thales Group, France • [12] Fraunhofer IAIS, Germany • [13] SAP, Germany • [14] Luxembourg Institute of Science and Technology, Luxembourg • [15] Institute of the Lithuanian Language, Lithuania • [16] University Pompeu Fabra, Spain • [17] University of Lisbon, Portugal • [18] University of Vienna, Austria • [19] University of Antwerp, Belgium • [20] University of Bergen, Norway • [21] Slovak Academy of Sciences, Slovakia • [22] Bulgarian Academy of Sciences, Bulgaria • [23] Jozef Stefan Institute, Slovenia • [24] University of Belgrade, Serbia • [25] University of Helsinki, Finland • [26] FBK, Italy • [27] Utrecht University, The Netherlands • [28] ICS, Polish Academy of Sciences, Poland • [29] University of Iceland, Iceland • [30] University of Malta, Malta • [31] University of Copenhagen, Denmark • [32] University of Latvia, Latvia • [33] University of Zagreb, Croatia • [34] Romanian Academy of Sciences, Romania • [35] MTA Research Institute for Linguistics, Hungary • [36] University of Tartu, Estonia • [37] Dublin City University and ADAPT Centre, Ireland • [38] CNRS, LIMSI, France

Corresponding author: Georg Rehm – georg.rehm@dfki.de

**Abstract**

Multilingualism is a cultural cornerstone of Europe and firmly anchored in the European treaties including full language equality. However, language barriers impacting business, cross-lingual and cross-cultural communication are still omnipresent. Language Technologies (LTs) are a powerful means to break down these barriers. While the last decade has seen various initiatives that created a multitude of approaches and technologies tailored to Europe's specific needs, there is still an immense level of fragmentation. At the same time, AI has become an increasingly important concept in the European Information and Communication Technology area. For a few years now, AI – including many opportunities, synergies but also misconceptions – has been overshadowing every other topic. We present an overview of the European LT landscape, describing funding programmes, activities, actions and challenges in the different countries with regard to LT, including the current state of play in industry and the LT market. We present a brief overview of the main LT-related activities on the EU level in the last ten years and develop strategic guidance with regard to four key dimensions.

## 1. Introduction

Europe has a long tradition of research in Language Technology (LT), which has not only enabled a highly visible and internationally recognised research community but also a large, diverse and growing LT industry. Commercial LT products have become indispensable in our day-to-day lives. The ability to communicate cross-lingually and, ultimately, cross-culturally using LT is crucial in Europe's multilingual society with its 24 official EU Member State languages and many more regional languages as well as languages of immigrants, minorities and trade partners.

The fragmentation of the European LT landscape is imposing severe challenges on the community, on various levels. The LT market, i. e., the commercial LT space in Europe, is extremely fragmented. There is a multitude of small providers, many of them addressing specific niches or verticals, who find it difficult to scale up, for example, by penetrating other, bigger markets or by competing on an international level with large or very large enterprises that have a competitive advantage in terms of research capacities, computing resources and data availability (Rehm, 2017; Vasiljevs et al., 2019). In addition, the META-NET White Paper Series has shown that there is a severe threat of digital extinction for at least 21 European languages because these languages are crucially under-resourced (Rehm and Uszkoreit, 2012; Rehm et al., 2014): for these, many types of technologies (including corpora) simply do not exist. This situation creates an urgent demand for new LT tailored to Europe's specific cultural, communicative and linguistic needs (Rehm and Uszkoreit, 2013; Rehm et al., 2016a).

Since 2010, the topic has been receiving more and more attention, recently also increasingly on a political level. In 2017, the study "Language Equality in the Digital Age – Towards a Human Language Project", commissioned by the European Parliament's Science and Technology Options Assessment Committee (STOA), concluded that the topics of LT and multilingualism are not adequately considered in current EU policies (STOA, 2017). Compared to other trending topics in the wider ICT area, LT has always had a somewhat "shadowy existence" and significantly less relevance. For example, the Digital Single Market Strategy of 2015 only touched briefly upon the need of multilingual services (European Commission, 2015).

Over the coming years, AI is expected to transform not only every industry but society as a whole. While other tasks such as image recognition and robotics have provided testbeds for massive new scientific breakthroughs, LT and NLP are, by now, considered important driving forces. An increasing number of researchers perceive full language understanding to be the next barrier and one of the ultimate goals of the next generation of innovative AI technologies (STOA, 2017). While AI has already achieved a lot of momentum, LT is catching up. Nevertheless, the European Parliament adopted, on 11 September 2018, with a landslide majority of 592 votes in favour, a resolution on "language equality in the digital age" that also includes the suggestion to intensify research and funding to achieve deep natural language understanding (European Parliament, 2018).

In this paper we provide an overview of the current European LT landscape from different perspectives. Colleagues from approx. 30 European countries have contributed short statements describing specific activities, funding programmes, actions and challenges in their respective countries with regard to LT (Section 2). Another important perspective to take into account is the current state of the European LT market (Section 3). We also examine the needs and demands of industry and research with regard to platforms and repositories (Section 4), as well as the evolution of the European LT landscape in the last ten years (Section 5). Evaluating all the different aspects, we can draw various conclusions and motivate next steps that the field, as a whole should take (Section 6).

## 2. The European LT Landscape

As part of the European Language Grid project (ELG; cf. Section 6.1) and based on the META-NET Network of Excellence, 32 ELG National Competence Centres (NCCs) were established. Following META-NET's work to assess the situation of the LT landscape in Europe (Rehm et al., 2016b), the NCCs were asked to provide information about corresponding funding programmes, activities and challenges in their countries (cf. Table 2 for a summary).

**Austria:** Currently, there is no specific funding programme for LT services, tools or resources. However, there is an initiative for digitalisation in industry, administration, and research. Recently, a new AI initiative was announced; the concrete funding programmes are yet to follow. While for German, in general, resources and LTs are widely available, very little is available for Austrian German.

**Belgium:** There is no specific funding programme, a lack of a joint programme with The Netherlands is also apparent. Belgium is currently re-applying for CLARIN membership and has limited funding for a CLARIAH programme in Flanders but the latter currently does not fund NLP research. Fragmentary research efforts on language-centric AI exist in National Science Foundation funding schemes.

**Bulgaria:** There is a need for a large collection of data sets and resources, services and tools for spoken language. The National Scientific Fund supports LT projects in common with all other disciplines. The "Innovation Strategy for Smart Specialization 2014–2020" identifies LT as a subdomain but there is no dedicated funding for Bulgarian LT yet.

**Croatia:** No funding programmes exist although there is a need for building up a new generation of LR/LT for Croatian. Some are developed within the cooperation of Croatian institutions as partners in COST, CEF and MSC projects. Connecting with neighbouring countries is expected as well as deeper involvement in CLARIN (through HR-CLARIN).

**Czech Republic:** There is no dedicated programme, but there has been some success for LT and AI; the total amount of funding in basic and applied research areas is over 2 million EUR p. a. The established research infrastructures, LINDAT/CLARIAH-CZ and the Czech National Corpus receive around 3.3 million EUR total p. a. The awareness of LT has increased significantly since the META-NET White Papers have been published; it is now listed as one of the three largest areas of research in the Czech government's National AI Strategy. Czech has now a relatively good coverage in terms of linguistically annotated resources.

**Denmark:** In 2018, the Ministry of Culture set up an LT Committee, led by the Danish Language Council. In March 2019, the Government presented The National Strategy for AI which includes an initiative of 4 million EUR for developing a Danish language resource to boost and scale up Danish language-centred AI. In April 2019, the recommendations of the LT Committee were published. The current status is as follows: The LT upgrade project is managed by The Danish Agency for Digitisation. Due to the change of government in June 2019, the project is still in an early stage, focus is on planning and organising. The aim is to develop a platform that contains free Danish LRs and functionalities aimed for the NLP industry. First steps include the upgrade of existing Danish dictionaries, and lexical resources as well as the development of a time-encoded Danish speech recognition corpus.

**Estonia:** There is a national programme for Estonian LT. Estonia is also prioritising AI and in 2019, an AI Action Plan was revealed. Applying LT in the public sector is one of the goals. More support for the Estonian language is needed at the user level (e. g., MT for all European and EU regulations, documents, discussions etc.). LT modules developed by Estonian researchers are useful and freely available but integration into end user applications or bigger human interface systems is rather poor.

**Finland:** The government has opened resources and databases produced by government-funded activities. In early 2019, The Ministry of Economic Affairs and Employment presented the final report from its "Finland leading the way into the age of AI" programme. In late 2019, the Ministry of Finance issued a "Development and implementation plan for AuroraAI 2019–2023", which includes the goal to identify service needs which the citizen expresses in natural language, written or spoken. The reports assume that LT is available for the languages used in Finland, so from 2019, the state-owned development company VAKE has included support for LT development in its strategy for digitalisation.

**France:** While there are no LT funding programmes, the government has launched an ambitious plan for research in all areas of AI, which has so far yielded the creation of four national AI institutes (in Paris, Toulouse, Grenoble and Nice), some of which (notably Toulouse and Grenoble, but also Paris) target LT in their roadmaps. A national programme of approx. 40 chairs and 200 PhDs in AI is currently under review, which should also include LT. This programme also targets international cooperation, first of all with Germany, Canada, and Japan. The main funding agency ANR funds three to five large-scale LT projects annually. The Ministry of Culture's small scale programme "Language and Digitalization" ran from 2016 to 2017.

**Germany:** While there is no LT-specific funding programme, the situation for LT research and development in Germany is rather good. In 2018, the government published its national AI strategy. Language analysis and understanding is (under the umbrella of HCI) one of five focus areas for innovation. The government aims to invest approx. 3 billion EUR until 2025 to implement the strategy, including the creation of new AI centres, new funding programmes, new professorships, new international collaborations (e. g., with France) and a new national roadmap for AI standardisation. For research and industry, these are additional opportunities on top of the established funding instruments (e. g., DFG, BMBF). It remains to be seen if LT projects will rather focus upon English (to be able to compete with the scientific community) or on German. The new project SPEAKER (2020-2023) is an example of the latter category, developing a conversational agent platform for the German language.

**Greece:** There is no LT-specific programme, but funding for LT, AI or language-related projects can be obtained through various programmes. LT and LRs are supported by the national project CLARIN:EL in the framework of the APOLLONIS infrastructure, with a total funding of 4 million EUR until 2020. Challenges consist in supporting lesser-resourced languages spoken in the country, namely, Greek but also languages of immigrants.

**Hungary:** There is no dedicated LT programme but some projects cover LT applications. Since 2012, at the Pázmány Péter Catholic University (PPKE) the independent Hungarian LT Research Group (MTA-PPKE) has been running with a funding of 70,000 EUR p. a., provided by the Hungarian Academy of Sciences, supporting the salaries of about six postdoc researchers and PhD students. Another opportunity is a recently started national AI research project. The focus is on neural methods and their applications in various areas, including LT. Its budget is approx. 3 million EUR, but LT comprises only a small part (approx. 100,000 EUR).

**Iceland:** The government decided to implement and fund a five-year project plan for Icelandic LT, starting in late 2019. The total budget is around 17 million EUR (13.5 from the government, 3.5 from industry). All resources and tools that will be developed with government money will be open and free. The challenge is to raise the interest of companies in developing and using LT tools and services.

**Ireland:** There is funding via Science Foundation Ireland (SFI). The lack of data is being improved by involvement in CEF projects ELRI and PRINCIPLE, which both have to do with data gathering and delivery to ELRC-SHARE. A national plan regarding Irish is in preparation. Its focus will be the provision of basic resources and technological building blocks for digital enablement. There is an emphasis on further linguistic study to enable the creation of richer data resources to drive solution development. In 2018, a budget of 1.66 million EUR was available for research projects.

**Italy:** The most recent LT funding programme dates back to 1999–2001. Since then, there has been no specific programme, nor is one foreseen in the near future. The National Programme for Research (2015–2020) identifies four technological clusters grouping 12 thematic areas – language is never mentioned. Italy lacks a coordinated plan for the development of LT. The Italian NLP sector finds some financial support by the national funding provided by the Ministry for University and Research (MIUR): in 2017, 111 million EUR out of a total budget of 391 million EUR were allocated to the wider SSH sector, where LT activities can receive some support (but LT is not recognised as a specific sector). On the regional level, programmes such as Working Regional Programme from the European Fund for Regional Development (2014–2020) reserve support to the wider ICT sector for the development of new technologies and innovation. Thus, funding support is, in principle, available depending on the initiative and capacity of individual researchers and groups, but Italy severely lacks a coordinated research and development framework.

**Latvia:** There is no LT funding programme, but some support exists from the governmental research programme "Latvian Language" and related projects (Skadiņa, 2019). It includes projects of the Latvian Council of Science (three projects develop a corpus of Latvian language acquisition, natural language understanding in HCI, and a Latvian WordNet), research and development projects supported through European Structural funds (two large projects are running currently – on HCI and on domain-specific speech recognition), as well as some funding from the project that supports research infrastructures (including CLARIN-LV). Funding in general remains a challenge.

**Lithuania:** Several LT policy documents and programmes exist in Lithuania. The "Guidelines for Lithuanian LT development 2014-2020", issued by the State Commission of the Lithuanian Language, are currently updated. Its main priorities were R&D for LT, MT, speech analysis, dialogue systems, summarisation, semantic technologies, text analysis, LRs, and others. "The Lithuanian Information Society Development Programme 2014–2020" promoted the Lithuanian culture and language through ICT by creating digital content based on text and speech interfaces, and developing digital products and services. The programme "The Lithuanian Language for Information Society (2014-2020)" was approved to ensure funding by European Structural funds and the State. Five projects develop LT services or resources, including speech, digital public services, machine translation, localisation. Recently, a Lithuanian AI strategy was prepared by the Ministry of the Economy and Innovation to ensure sustainable development of AI.

**Luxembourg:** There is no specific LT programme. In 1984, Luxembourgian (or Luxembourgish) was "officialised" as a national language. At the University of Luxembourg, there is the Institut de langue et de littératures luxembourgeoises (Institute of Luxembourgish language and literature), which had, until 2014, projects on lexicography and phraseology. The latest project (2013-2016) is about the standardization of the German language in Luxembourg. There have also been some projects related to LT funded by INTERREG Grande Région. For 2014-2020, INTERREG Grande Région has a budget of 140 million EUR. There is no dedicated funding for LT. The Luxembourg National Research Fund has recently reviewed its priorities. One of these is the multilingual situation in the school system.

**Malta:** LT for Maltese has a chequered history due to the lack of a systematic funding programme. Nevertheless there have been a number of small-scale projects during the last 20 years, funded from national and EU sources including the government. These have produced the Maltese Language Resource Server, which now includes large-scale corpora of Maltese text, learner corpora of Maltese English and Maltese, lexicons, and some NLP tools. A speech synthesiser was developed in 2012 with ERDF support. In 2018, the University of Malta research fund has awarded 50,000 EUR for the joint development of a speech recognition software by the Institute of Linguistics and LT and the Department of AI. The government has allocated funds for the future development of a spellchecker, and further funding is expected for implementing the recent Malta AI Strategy (Oct. 2019) which states that the country will make crucial investments in the development of Maltese LRs and tools enabling computers to process, understand and generate Maltese text and speech, and develop AI services in both of Malta's official languages.

**The Netherlands:** There is no dedicated programme for LT development, though several projects are ongoing. Some LT development takes place in the context of CLARIAH, especially on speech recognition, event extraction and POS tagging. There is, thanks to the STEVIN programme (2004-2009), no immediate danger for digital extinction of the Dutch language. The META-NET White Papers increased the awareness of the Interparliamentary Committee for the Dutch Language Union of the importance of LT. In 2015, without committing any funding the committee invited the Dutch LT community to submit a proposal for a new LT programme. Such a proposal was never defined. The AI Research Agenda, recently published by NWO, includes NLP.

**Norway:** There is no funding programme for LT. The preparatory research programme (KUNSTI) was highly successful but has not been followed up. Norway has established an LT resource collection for Norwegian (Språkbanken), which shares open resources for research and development of LT products and services.

**Poland:** No specific programme exists but certain support for LT is directed at CLARIN-PL through funding the national contribution in a programme by the Ministry of Science and Higher Education dedicated to key research infrastructures. The main challenge remains the further development of the National Corpus of Polish, a resource with enormous impact on research in linguistics, humanities and LT, which was completed in 2011 and not been updated since. LT is mentioned in the "Policy for the Development of AI in Poland for the years 2019–2027" as a key function of AI but no specific actions are planned. In 2020 another infrastructure with LT components, DARIAH-PL, joined CLARIN-PL on the Polish Map of Research Infrastructure (PMIB).

**Portugal:** Currently, there is no specific funding programme for LT. However, there is a general initiative to foster digital transformation in industry, administration, and research. A new AI initiative has been announced; the concrete funding schemes and programmes are yet to follow.

**Romania:** There is no LT-specific funding program. The LT R&D (mainly speech) is embedded into AI proposals. Currently two large projects are active: "Robots and the Society: Cognitive Systems for Personal Robots and Autonomous Vehicles" (ROBIN) and "Resources and Technologies for the development of man-machine interfaces for Romanian language" (ReTeRom). Both have a significant part related to speech LRs and processing of Romanian.

**Serbia:** There are no LT funding programmes. Recently, the government has established a working group with the aim of formulating a strategy for the development of AI 2020-2025. As of yet, it is unclear if LT will have a specific place in it. Also, an AI institute has just been established in the Science Technology Park with the aim of connecting academia and industry. In 2020 thoroughly reorganized funding of national scientific programs was initiated by the first call for proposals in the field of AI, including LT. The results of this call are not known yet. Industry (in the broadest sense) is a modest user of existing LT, and even lesser is their involvement in the development of LT for Serbian.

**Slovakia:** There is no LT funding programme; some minor programmes oriented towards LT have been successful in the past, but mostly as parts of other actions or grants. LT-oriented industry is rare, with companies usually trying to use existing technologies rather than developing new ones. There is a lack of understanding of NLP within the industry.

**Slovenia:** There is only funding for CLARIN.SI (approx. 100,000 EUR p. a.), and the long-term research programme "LRs/LTs for Slovene" (2,5 FTE). A 4 million EUR call was published in Nov. 2019 covering LT topics for Slovene such as speech technology, MT, semantic technologies, corpus upgrades and a terminology portal.

**Spain:** The Plan for the Promotion of LT was approved in 2015 to promote the development of NLP, automatic translation and conversational systems in Spanish and co-official languages in areas like health, justice, and technology watch. It has focused on the production of resources and basic tools for Spanish and other languages in Spain.

**Sweden:** There is no LT funding programme. LT project proposals are evaluated as and competing with proposals in computer science or linguistics. An AI funding programme is being planned, which will benefit at least some forms of LT. The Swedish Research Council is currently providing substantial funding to a national research infrastructure supporting research based on language data, notably LT, linguistics and digital humanities, for 2018–2024.

**United Kingdom:** The EPSRC funds NLP actions, but this is primarily blue skies research. It is currently designated as a growth area, meaning that it is actively tried to increase the funding allocated to NLP research. However, LT and research infrastructures, in general, are not perceived as high funding priority (e. g., the UK is only an observing member of CLARIN). At the same time, GATE and, most recently, GATE Cloud are among the most widely used and established LT tools, services, and platforms.

## 3. Language Technology Market

The final study report on the value proposition of CEF eTranslation (Section 5) in the context of the European LT market and ecosystem was commissioned by the EU and published in 2019 (Vasiljevs et al., 2019). It positions CEF eTranslation, one of the infrastructural building blocks of the Connecting Europe Facility, in the European LT market. The study provides an analysis of the EU market, of LT adoption by public services in the EU, and of the EU's competitiveness with respect to the US and Asia in three areas (i. e., MT, speech technology, cross-lingual search). The study develops a business model for CEF eTranslation by defining its value proposition in the context of the market. The analysis has the objective of providing an overview of the EU LT market together with the emerging trends and an estimate of its growth. An exhaustive list of LT companies active in EU Member States was created; 473 of those fully qualify as LT vendors. Based on desk research using public sources and in-house databases of consortium member IDC, the total size of the LT industry within the EU26 plus Iceland and Norway in 2017 was estimated at approx. 800 million EUR, which is a relatively small market in IT terms. Germany holds the largest share, followed by the UK. Forecasts predict the market to grow at an average rate of 10% between now and 2021.[1]

|  | 2018 | 2019 | 2020 |
|---|---|---|---|
| Germany | 197M€ | 217M€ | 240M€ |
| United Kingdom | 189M€ | 209M€ | 232M€ |
| France | 88M€ | 96M€ | 105M€ |
| Netherlands | 55M€ | 60M€ | 66M€ |
| Rest of EU 28 | 249M€ | 277M€ | 305M€ |
| **Total** | **778M€** | **859M€** | **948M€** |

Table 1: Size of the European LT market (forecast 2018-2020; numbers taken from Vasiljevs et al., 2019)

This desk research was complemented by primary research which consisted of analysing the responses of a questionnaire and information provided through phone interviews. Based on 51 questionnaire responses and eight subsequent interviews, the consortium was able to get a picture of the market size, language offering, types of LT offered, customer segments, and perception of the future. The LT market in Europe is very fragmented and composed of small and medium-sized enterprises (SMEs), typically local players providing local solutions. Profitability is quite low, competition intense and margins compressed. One of the reasons for this low vendor profitability is the need for continuous innovation and the cost related to this need. In terms of language offering, English, German, French, Spanish and Italian are of utmost importance to the LT vendors. As LT markets for most European languages are small, business opportunities are limited for vendors that focus on particular languages. In terms of the types of LT offered, translation technology is considered the biggest revenue contributor followed by speech technology. Multilingual and search technology are least important in terms of revenue.

As for customer segments, vendors consider the public sector the most important segment, though it accounts for only 20% of their revenues and lags behind the private sector in terms of profitability. Most suppliers are quite positive when looking towards the future and expect the LT market to grow, as AI will increasingly be an integral part of LT. Natural Language Understanding (NLU) in general and chatbot applications in particular were often mentioned as emerging technologies to look out for and are expected to become increasingly widespread (Vasiljevs et al., 2019).

## 4. Demands and Needs of European LT Industry and Research

In the summer of 2019 a survey was carried out in the scope of the ELG project (cf. Section 6.1) to assess the experience and needs of LT researchers, developers and professionals with regard to LT infrastructures (Melnika et al., 2019). The survey was distributed to key LT communities (META-NET, CRACKER, LT-Innovate, ELRA, CLARIN, ELRC and others). There were 158 responses, 50 of which were incomplete, resulting in 108 complete responses included in the analysis. The largest group of respondents were from academic and research institutions (67%), followed by industry players (24%); the rest represented public administrations, NGOs and freelancers. The aim of the survey was to explore user experience and gather feedback on existing infrastructures, as well as to determine the factors influencing significant decisions of users.

Two thirds of the respondents search for LRs in various repositories, about 65% use a generic search engine (e. g., Google Search). When searching for language data, they mostly search in the CLARIN repository (64%), the catalogues of ELRA (56%) or LDC (56%), generic software repositories like GitHub, GitLab (56%) or META-SHARE (44%). When looking for language processing software tools, generic software repositories are the first place they look (71%), followed closely by generic search engines (69%), the CLARIN repository (42%), institutional repositories (35%), or META-SHARE (29%).

Key criteria in selecting language processing software are language coverage (62%), licensing conditions (59%), the availability of open source code (56%), usability (46%), performance (42%) and the availability of documentation (41%). Although most respondents prefer to download binaries or source code (72%), there is also a strong interest to download containerized tools (63%). Using services through an available web interface is also important (39%). Although the majority of respondents (61%) have provided language data or software/tools to an online platform, only 46% of industry players have done so. Specific institutional repositories (73%) and generic online software repositories, e. g., GitHub or GitLab (64%) are used the most for submitting language data, followed by the CLARIN repository (48%), META-SHARE (33%) and the ELRA catalogue (23%). For software sharing, GitHub, GitLab and docker are used the most (77%), followed by institutional platforms (62%), CLARIN (36%), and META-SHARE (21%).

---
[1] A more recent forecast estimates the global NLP market to reach $29.5B by 2025, see https://www.reportlinker.com/p05838704/Global-Natural-Language-Processing-Market.html.

Promotion of their LRs (71%) and goodwill (64%) are the primary motivations for sharing. Ensuring reproducibility of research results (55%), project requirements (42%), and promotion of the organization (21%) are other motivating factors. Major obstacles in sharing are copyright/IPR issues and a lack of clarity about the type of license to assign (42% and 35% respectively).

It is notable that 70% of respondents are interested in providing their software as containers and only 12% are not. Respondents have noted that when providing software or language data, they prefer the process to be quick and easy (97%), access should be secure (74%).

When asked about future intentions, only 2% do not anticipate to contribute any kind of language content. On the other hand, all respondents indicated that they would be interested in continuing to search for and use the various types of content (tools and services – 89%, data – 88% etc).

Although many LR repositories exist, many respondents can neither find the necessary data/software (44%), nor samples or demo versions (35%), find licensing and usage conditions to be unclear (33%), get too many irrelevant results (32%) or find descriptions and metadata insufficient (31%). Fragmentation is also an issue. This is why, for 66% of respondents, it is important or very important to have one centralized digital meeting spot for LTs in Europe.

## 5. Recent Developments (2010-2020)

Starting with META-NET in 2010, a substantial number of initiatives and projects have attempted to foster research, innovation and development towards a truly multilingual Europe, enabled and supported by LT "made in Europe". Below, we list a selection of the relevant activities, concentrating on LT first (Section 5.1) and AI second (Section 5.2).

### 5.1. LT-specific Initiatives

**META-NET** Founded in 2010, META-NET[2] is a European Network of Excellence dedicated to the technological foundations of a multilingual and inclusive European society, bringing together 60 research centres in 34 European countries. META-NET was, between 2010 and 2017, supported through the EU projects T4ME, CESAR, METANET4U, META-NORD and CRACKER. One of its main goals is technology support for all European languages as well as fostering innovative research by providing strategic recommendations with regard to key research topics (Rehm and Uszkoreit, 2013). META-SHARE[3] is an infrastructure that brings together providers and consumers of language data, tools and services. It is a network of repositories that store resources, documented with high-quality metadata aggregated in central inventories (Piperidis, 2012; Gavrilidou et al., 2012; Piperidis et al., 2014).

**CLARIN ERIC** The CLARIN European Research Infrastructure for Language Resources and Technology is a legal entity set up in 2012, with 20 member countries at present.[4] CLARIN makes language resources available to scholars, researchers, and students from all disciplines with a focus on the humanities and social sciences. CLARIN offers solutions and services for deploying, connecting, analyzing and sustaining digital language data and tools.

**Call ICT-17-2014 – "Cracking the Language Barrier"** The EU call ICT-17-2014, which was informed by key META-NET results (Rehm and Uszkoreit, 2012), funded a total of six projects including, among others, QT21[5], Modern MT (MMT)[6], Health in My Language (HimL)[7] and the Coordination and Support Action CRACKER[8]. CRACKER initiated the Cracking the Language Barrier[9] federation and continued META-SHARE maintenance. The federation was established as an umbrella initiative, which brought together more than 20 relevant organisations and projects working on technologies for multilingual Europe (Riga Declaration, 2015; Rehm et al., 2016a), emphasising the idea of an initiative *from* the community *for* the community.

**CEF eTranslation and ELRC** The EC has been making use of LT internally for several years. CEF eTranslation is an automated translation platform that supports the multilinguality of public services, in particular CEF Digital Service Infrastructures, in Europe. The system builds upon MT@EC, the EC's MT service that was internally deployed in 2013 and which is available to EU institutions, Member State administrations and various EC information systems and online services. The eTranslation system is supported through a number of service contracts such as, crucially, European Language Resource Coordination (ELRC), that provides reach into the Member States and that promotes the collection and sharing of language data and their ingestion into the eTranslation system (Lösch et al., 2018).[10]

**STOA Workshop and Study** After a first European Parliament workshop on MT in December 2013, the workshop "Language equality in the Digital Age – Towards a Human Language Project" took place in the EP in January 2017.[11] The discussion circled around the idea of a Human Language Project, a large-scale, long-term flagship initiative to carry out research, development and innovation activities. In this project, new breakthroughs towards Deep Natural Language Understanding were supposed to be made to address the threat of digital language extinction and also to provide solutions to the European citizens, industry and administrations. After the workshop, a report was published, commissioned by the EP (STOA, 2017). This detailed study provides 11 policy recommendations towards the European Institutions. In the wake of the STOA report, the HLP Prep consortium applied (unsuccessfully) for one of the six EU FET Flagship preparation projects in 2018.[12]

**Call ICT-29-2018 – "A Multilingual Next Generation Internet"** Most recently, LT, as a broader topic, was re-included in the Horizon 2020 Work Programme 2018-20 under the call ICT-29-2018 with a budget of 25 million

---

[2] http://www.meta-net.eu
[3] http://www.meta-share.org
[4] https://www.clarin.eu
[5] http://www.qt21.eu
[6] https://www.modernmt.com
[7] http://www.himl.eu
[8] http://cracker-project.eu
[9] http://www.cracking-the-language-barrier.eu
[10] http://www.lr-coordination.eu
[11] http://www.stoa.europarl.europa.eu/stoa/cms/home
[12] http://human-language-project.eu

EUR. The call resulted in six research projects – Bergamot[13] (client-side MT in the browser), COMPRISE[14] (multilingual, privacy-driven voice-enabled services), ELITR[15] (European live translator), EMBEDDIA[16] (cross-lingual embeddings for less-represented languages in news media), gourmet[17] (global under-resourced media translation), Prêt-à-LLOD[18] (multilingual linked language data for knowledge services across sectors) – and the Innovation Action European Language Grid (ELG).[19]

**EP Resolution "Language Equality in the Digital Age"**
In a plenary meeting on 11 September 2018, the European Parliament adopted this joint ITRE/CULT report (European Parliament, 2018) with a majority of 592 votes in favour, 45 against and 44 abstentions. The adoption of the resolution with a "landslide" majority demonstrates the importance and relevance of the topic. The ten-page resolution includes 45 recommendations. An informal survey among the ELG National Competence Centres selected the following four recommendations as the most important ones:

1. "[…] owing to a lack of adequate policies in Europe, there is currently a widening technology gap between well-resourced languages and less-resourced languages […]; […] more than 20 European languages are in danger of digital language extinction; notes that the EU and its institutions have a duty to enhance, promote and uphold linguistic diversity in Europe."
28. "[…] specific programmes within current […] as well as successor funding programmes, should boost long-term basic research as well as knowledge and technology transfer between countries and regions."
32. "Urges the Commission to set up an HLT financing platform, drawing on the implementation of [FP7, Horizon 2020, CEF]; […] the Commission should place emphasis on research areas needed to ensure a deep language understanding, such as computational linguistics, linguistics, artificial intelligence, LTs, computer science and cognitive science."
45. "[…] stresses the need to adapt the regulatory framework and ensure a more open and interoperable use and collection of language resources; […]"

### 5.2. AI-specific Initiatives

**CLAIRE** The Confederation of Laboratories for AI Research in Europe (CLAIRE) was launched on 18 June 2018 with a vision document signed by 600 senior researchers and key stakeholders in AI.[20] That document calls for "Excellence across *All of AI*. For all of Europe. With a Human-Centred Focus." CLAIRE has since garnered support for that vision by more than 3,300 individuals from Europe, with about two thirds being AI experts. Additionally, the world's largest network of AI labs across Europe has been created with more than 350 labs, representing over 20,000 employees. CLAIRE is now supported by nine EU Member State governments. CLAIRE collaborates closely with HumanE-AI, AI4EU, and plays a crucial coordinating role in many of the "AI excellence centres" proposals (ICT-48-2020). CLAIRE is deeply involved in the preparations of the AI PPP (see below). One of the nine CLAIRE Informal Advisory Groups (IAGs) focuses upon NLP. Two closely related IAGs are Machine Learning (ML) and Knowledge Representation and Reasoning (KRR).

**HumanE AI** The EU project Human-Centered Artificial Intelligence (HumanE AI) is one of six EU FET Flagship preparation projects.[21] Its goal is to design AI systems that enhance human capabilities and empower individuals and society as a whole to develop AI that extends rather than replaces human intelligence. This vision fits very well into the ambitions articulated by the EC but cannot be achieved by legislation or political directives alone. Instead, it needs fundamentally new solutions to core research problems in AI, especially to help people understand actions recommended or performed by AI systems. Among the key challenges are achieving in-depth understanding of humans and social contexts, including human language.

**AI4EU** The European AI On Demand Platform (AI4EU) is a EU-funded project in which 81 partners from 21 countries develop a platform that bundles and connects the European AI resources with the goal of ensuring European independence and leadership in AI research and innovation.[22] The platform is planned to act as a broker, developer and one-stop shop providing services, expertise, data, computing resources, among others. The initiative plans to build the platform upon and make it interoperable with existing AI and data components as well as platforms.

**PPP on AI, Data, Robotics** The two existing industry-driven associations BDVA[23] (Big Data Value Association) and euRobotics[24] have recently joined forces and started a collaboration towards the vision of establishing a PPP on AI, Data and Robotics. The idea is grounded in combining their core networks and making use of strong industrial as well as scientific ties in order to give AI the power to transform the economy as well as society, while preserving European values, with a focus on industrial environments and applications. The joint vision and strategic roadmap (Zillner et al., 2019) emphasises the aspect of open collaboration with different initiatives, including industry and academia. Most recently, representatives from the European AI community (incl. CLAIRE, ELLIS, EurAI, HumanE AI and AI4EU) have started contributing to the PPP draft documents.

## 6. Conclusions – Future Work – Next Steps

We conclude by emphasising and highlighting four crucial areas of current and future work, i.e., a shared platform for the European LT landscape (Section 6.1), the aspect of platform interoperability (Section 6.2), establishing a representation of the European LT community (Section 6.3) and language-centric AI (Section 6.4).

---

[13] https://browser.mt
[14] https://www.compriseh2020.eu
[15] https://elitr.eu
[16] http://embeddia.eu
[17] https://gourmet-project.eu
[18] http://www.pret-a-llod.eu
[19] https://www.european-language-grid.eu
[20] https://claire-ai.org
[21] https://www.humane-ai.eu
[22] https://www.ai4eu.eu
[23] http://www.bdva.eu
[24] https://www.eu-robotics.net

## 6.1. Towards a Shared LT Platform

Multilingualism is at the heart of the European idea and one of its greatest assets of cultural diversity. The principle that all 24 official Member State languages have the same status is perpetuated in the EU Charter as well as in the Treaty on the EU. As also emphasised by the STOA Report and EP Resolution, there is a big need for a shared platform that bundles repositories and applications to benefit European society, industry and politics. The European Language Grid (ELG) project is currently developing a platform and joint market place that is meant to address this need and the fragmentation of the European LT landscape by providing access to LT services and data sets (Rehm et al., 2020a; Labropoulou et al., 2020). This scalable cloud platform will ultimately provide access to hundreds of commercial and non-commercial LTs for all European languages in an easy-to-integrate way, including running services, tools, data sets and resources. It will enable the European LT community to deposit and upload their technologies and data, to deploy them through the grid, and to connect them with other resources. The ELG will boost the Multilingual Digital Single Market towards a thriving European LT sector, creating new innovations, new jobs and new opportunities. Starting in 2020, ELG will begin to close at least some open gaps in terms of missing data sets or technologies through its two open calls with 15-20 pilot projects.

## 6.2. Towards Platform Interoperability

In addition to several existing ones, the EU and many European countries (in national projects) as well as companies currently develop new cloud platforms for the wider AI as well as LT landscape, including AI4EU and ELG but also various commercial cloud platforms. If these platforms are unable to communicate with each other, there is a danger that this proliferation will contribute to, rather than reduce, the ubiquitous fragmentation of the European landscape. All stakeholders must make an effort to collaborate on the interoperability of our existing and emerging platforms (Rehm et al., 2020b). AI4EU and ELG have already initiated a collaboration on several levels that will include, among others, automated mechanisms between the two platforms that allow the exchange of metadata records that provide structured and semantically aligned information about the contents of the respective platforms. One of the goals is also to share services, models and data sets.

## 6.3. Towards a European Representation

One crucial aspect that has been impacting the LT area's visibility on the international level is the lack of a complete European representation of the field. Previous approaches have been selective and, thus, incomplete. They either did not include all countries and their respective languages or they only addressed certain verticals or niches of the LT field, which is, by nature, extremely broad. Embracing this diverse "Multilingual Europe" community is an important component of the overall ELG concept, which includes LT provider companies, LT user/buyer companies, research centres and universities involved in LT research, development and innovation activities, the language communities, politics and public administrations, funding agencies, language service providers and translators as well as European citizens. In addition to the National Competence Centres (NCCs)[25], which are meant to support the ELG project itself, ELG also initiates a new body, the European LT Council (LTC), which is meant to represent the whole European LT field including all stakeholders, industry sectors, countries and related pan-European initiatives (e. g., CLAIRE, AI4EU, AI PPP, CLARIN, LT-Innovate etc.); the LTC will be fairly large, eventually consisting of approx. 200-225 members. It takes on strategic tasks and is expected to support the idea of technology-enabled and technology-supported multilingualism in Europe in general. The LTC will enable easy and efficient communication and coordination on the European level, specifically with regard to ongoing and emerging international activities related to LT research, development and innovation. It will foster the coordination and strategic as well as political discussion including the preparation of strategic recommendations, especially geared towards national and European administrations and funding agencies

## 6.4. Towards Support for Language-centric AI

As emphasised by the STOA Report and EP Resolution, there is an enormous need for a large-scale, multidisciplinary LT development and deployment programme that benefits European society, industry and politics. The opportunities of developing technologies for cross-cultural communication in Europe, and beyond, are almost endless. Both the scientific and technological roots of LT are deeply embedded in AI and Computational Linguistics, especially with regard to the development of knowledge-based systems for Natural Language Understanding. Today, there is no clear separation between LT and AI anymore, the boundary is becoming more and more blurred. Given the ubiquity of relevant frameworks, methods and datasets for the standard tasks and challenges, it may, thus, make more sense to frame the field as *Language-Centric AI* rather than *Language Technology*. In fact, language-centric AI is already scientific reality because many language-related technology approaches make use of deep learning frameworks that are typically more associated with AI in general than LT. Deep Natural Language Understanding can only be achieved by taking additional modalities and contexts into account, grounding utterances and discourse in communicative scenarios. Corresponding research must be interdisciplinary and necessarily include expertise from bordering areas. The other way around, AI research that concentrates on ambient spaces or human-machine interaction must also include the language modality, obviously, as already stressed in HumanE AI, among others. It is safe to assume that the four "AI excellence centres", to be funded through the call ICT-48-2020, will emphasise the interdisciplinary collaboration in AI to tackle not only the big societal challenges but also ethical issues, bias, trustworthy and explainable AI etc. Language-centric AI is an integral part of the AI community and will contribute to the further shaping of the European way of carrying out AI research, especially under the umbrella of the upcoming funding instruments Horizon Europe and Digital Europe Programme.

---

[25]https://www.european-language-grid.eu/ncc/

## 7. Acknowledgments

The authors would like to thank the three anonymous reviewers for their insightful comments.

The work presented in this paper has received funding from the European Union's Horizon 2020 research and innovation programme under grant agreements no. 825627 (European Language Grid, ELG), no. 761758 (Human-Centered Artificial Intelligence, HumanE AI), no. 825619 (A European AI On Demand Platform and Ecosystem, AI4EU), no. 732630 (Big Data Value ecosystem, BDVe), from the MEYS, Czech Republic, Project LM2018101 LINDAT/CLARIAH-CZ, and also from FCT, Portugal, Project PINFRA/22117/2016 PORTULANCLARIN. The ADAPT Centre for Digital Content Technology is funded under the SFI Research Centres Programme (Grant 13/RC/2106). The paper is, furthermore, supported by CLAIRE (Confederation of Laboratories for Artificial Intelligence Research in Europe) and ELRC (European Language Resource Coordination).

# Appendix

|  | LT-related funding | | | Artificial Intelligence | |
|---|---|---|---|---|---|
|  | None at all | Some funding | LT programme | AI strategy | LT funding through AI |
| Austria | X |  |  | X |  |
| Belgium |  | X |  |  | X |
| Bulgaria |  | X |  |  |  |
| Croatia | X |  |  |  |  |
| Czech Republic |  | X |  | X |  |
| Denmark |  |  | X | X | X |
| Estonia |  |  | X | X | X |
| Finland |  | X |  | X |  |
| France |  | X |  | X | X |
| Germany |  | X |  | X | X |
| Greece |  | X |  |  |  |
| Hungary |  | X |  | X |  |
| Iceland |  |  | X |  |  |
| Ireland |  | X |  |  |  |
| Italy |  | X |  | X |  |
| Latvia |  | X |  |  |  |
| Lithuania |  | X |  | X |  |
| Luxembourg |  | X |  |  |  |
| Malta |  | X |  | X | X |
| The Netherlands |  | X |  | X |  |
| Norway |  | X |  |  |  |
| Poland |  | X |  | X |  |
| Portugal |  | X |  | X |  |
| Romania |  | X |  |  |  |
| Serbia | X |  |  | X |  |
| Slovakia | X |  |  |  |  |
| Slovenia |  | X |  | X |  |
| Spain |  |  | X | X |  |
| Sweden |  | X |  | X |  |
| UK |  | X |  | X |  |
| Perc. | 13.3% | 73.3% | 13.3% | 63.3% | 20.0% |
| *European Union* |  | X |  | X | ? |

Table 2: Overview of the Language Technology funding situation in Europe (2019/2020)